\documentclass[letterpaper, 10 pt, conference]{ieeeconf}

\IEEEoverridecommandlockouts                             
\title{\LARGE \bf
Dynamic Registration: Joint Ego Motion Estimation and 3D Moving Object Detection in Dynamic Environment
}

\author{Wenyu Li$^{1}$ , Xinyu Zhang$^{2,\ast}$ , Zijun Wang$^{1}$ , Shichun Guo$^{2}$ , Nan Qiu$^{2}$ and Jun Li$^{2}$% <-this % stops a space
\thanks{$^{1}$Wenyu Li and Zijun Wang with with the School of Aeronautics and Astronautics,
	University of Electronic Science and Technology of China, Chengdu
	610054, China
        {\tt(202022100534@std.uestc.edu.cn;
        	wangzijun@uestc.edu.cn)}}%
\thanks{$^{2}$Xinyu Zhang, Shichun Guo, Nan Qiu and Jun Li are with Tsinghua University, Beijing, 100084 China.
        {\tt(xyzhang@tsinghua.edu.cn;
        	shichunguo@gmail.com;
        	nanq@mail.tsinghua.edu.cn;
        	lijun1958@tsinghua.edu.cn))}}%
\thanks{$^{\ast}$Author to whom correspondence should be addressed. (e-mail:
	{\tt(xyzhang@tsinghua.edu.cn)}.}
}

\usepackage[ruled,linesnumbered]{algorithm2e}
\usepackage{graphicx}
\usepackage{subfigure}
\graphicspath{{./figure/}}
\DeclareGraphicsExtensions{.pdf,.jpeg,.jpg,.png,.jpg,.eps}

\usepackage[font=small,tableposition=top,hypcap=false]{caption}

\usepackage{amsmath}
\usepackage{amssymb}
\usepackage{multirow}
\usepackage{booktabs}

\usepackage{cite}
\makeatletter
\let\NAT@parse\undefined
\makeatother
\usepackage[colorlinks,linkcolor=black,citecolor=green]{hyperref}
\usepackage[capitalize]{cleveref}

\begin{document}

\maketitle
\thispagestyle{empty}
\pagestyle{empty}

\begin{abstract}
	
Localization in a dynamic environment suffers from moving objects. Removing dynamic object is crucial in this situation but become tricky when ego-motion is coupled. 
In this paper, instead of proposing a new slam framework, we aim at a more general strategy for a localization scenario. In that case, Dynamic Registration is available for integrating with any lidar slam system.
We utilize 3D object detection to obtain potential moving objects and remove them temporarily.  Then we proposed Dynamic Registration, to iteratively estimate ego-motion and segment moving objects until no static object generates. Static objects are merged with the environment. Finally, we successfully segment dynamic objects, static environments with static objects, and ego-motion estimation in a dynamic environment.
We evaluate the performance of our proposed method on KITTI Tracking datasets. Results show stable and consistent improvements based on other classical registration algorithms.
	
\end{abstract}

\section{Introduction}
Simultaneous Localization and Mapping (SLAM) estimating the robot pose while mapping its surroundings of an unknown environment. 
Most SLAM systems assume static environments. However, the real world contains dynamic objects. Localization or ego-motion estimation in a dynamic environment is important for applications such as robot navigation and autonomous driving tasks.
SLAM algorithms can easily fail in dynamic environments, so moving objects need to be detected and removed.

Most articles \cite{saputra2018visual,bescos2021dynaslam, zhang2020vdoslam} viewed localization in a dynamic environment from two different perspectives: either as a robust SLAM  or as an extension of SLAM. The former regards dynamic elements as outliers and remove them from the estimated process \cite{mur-artal2017orbslam2}, whereas the latter extends SLAM to detect and track the dynamic objects \cite{wang2003online}. 

Robot systems require object information of the surroundings for decision making and motion planning.  In autonomous driving scenarios, the car not only needs to localize itself but to perceive other cars or pedestrians. 
Dynamic scenes are crucial in real environments. Thus it's necessary to pay attention to all objects and their motion state, to facilitate segmenting and tracking these instances, rather than focusing on segmenting and inferencing points that are probably in moving.

\begin{figure}[htbp]
	\centering
	\subfigure[3D Detection]{\label{fig:notremove}
		\includegraphics[width=0.4\columnwidth]{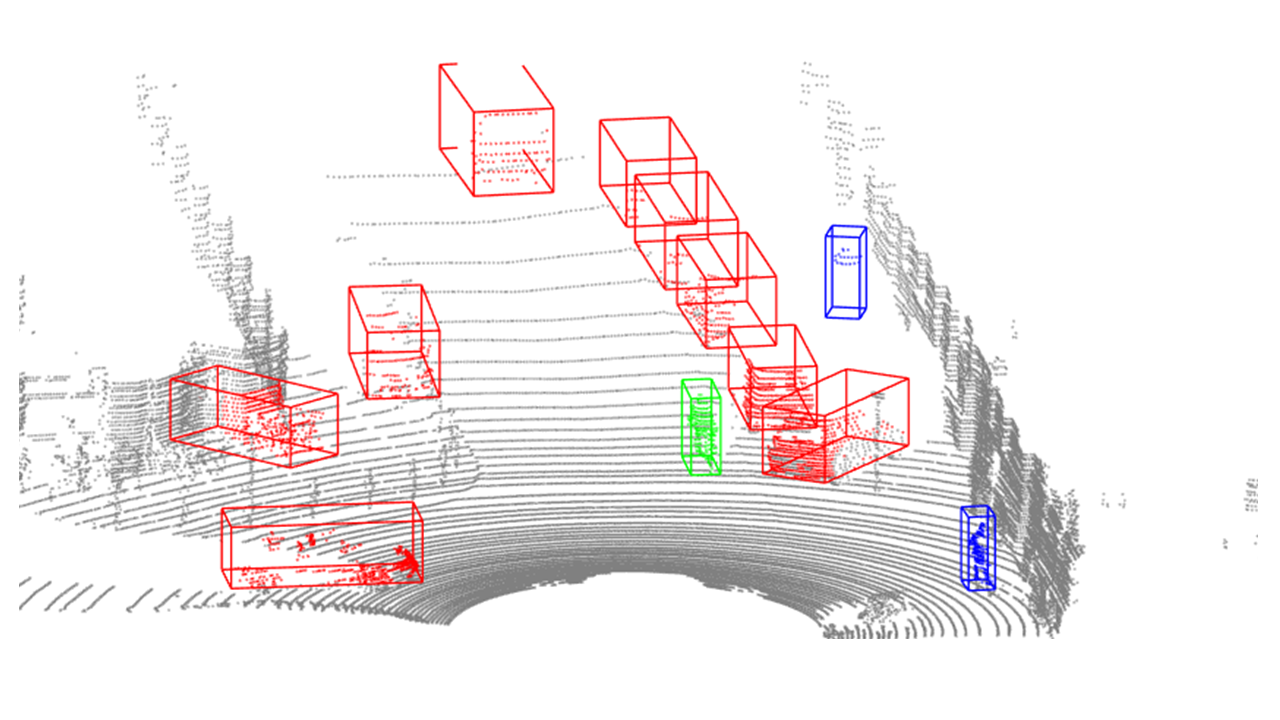}}
	\subfigure[Remove all objects]
	{\label{fig:removeall}
		\includegraphics[width=0.4\columnwidth]{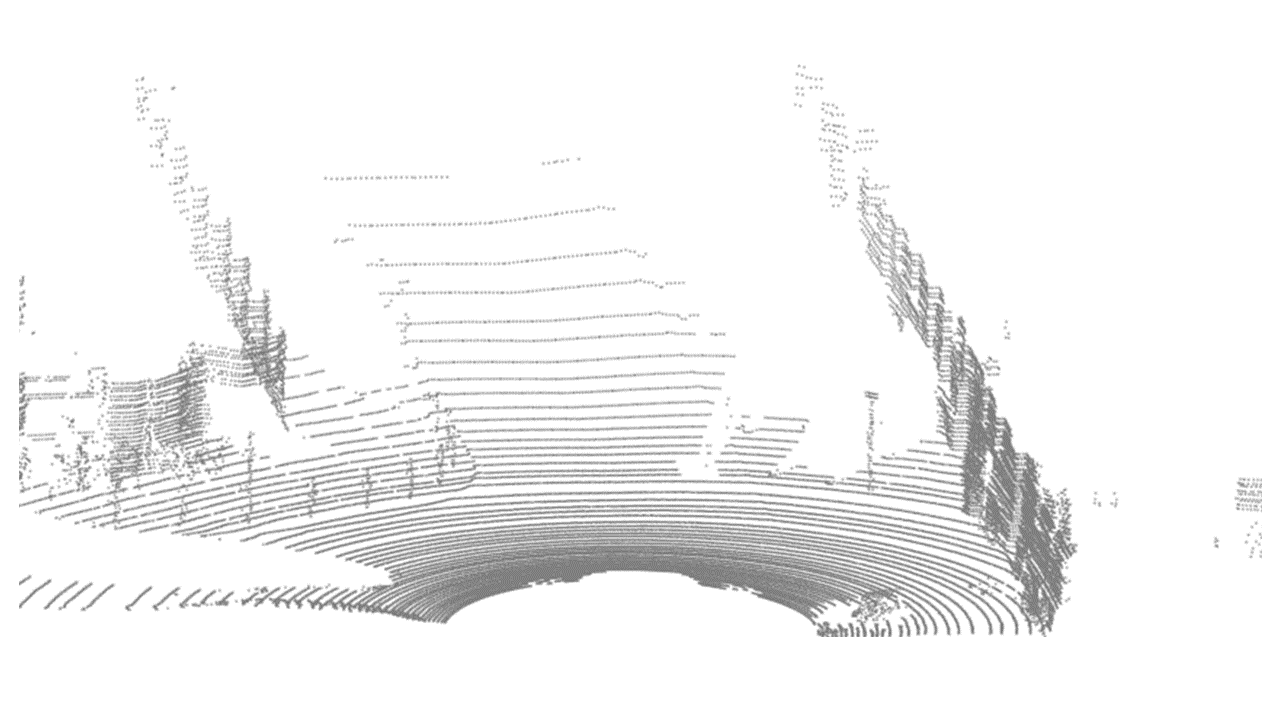}}
	\subfigure[Remove moving objects]{
		\label{fig:removedynamic}\includegraphics[width=0.4\columnwidth]{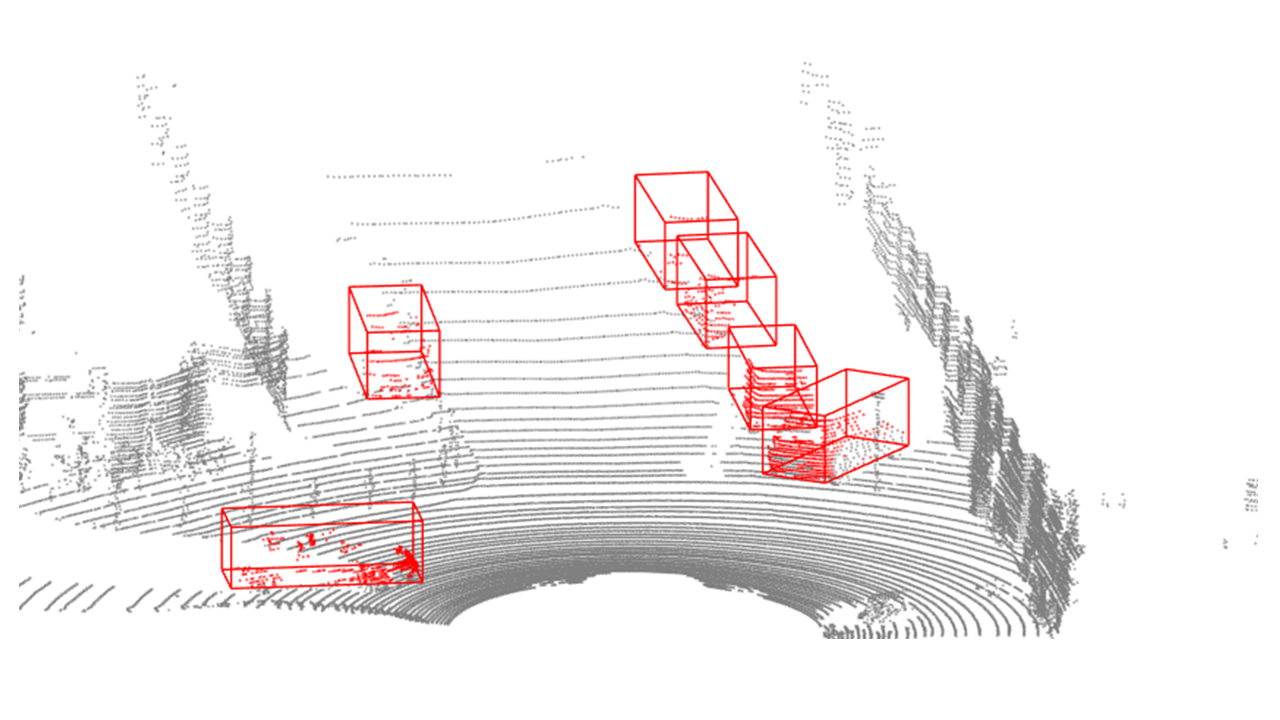}}
	\subfigure[Corresponding image ]{\label{fig:corimage}
		\includegraphics[width=0.4\columnwidth]{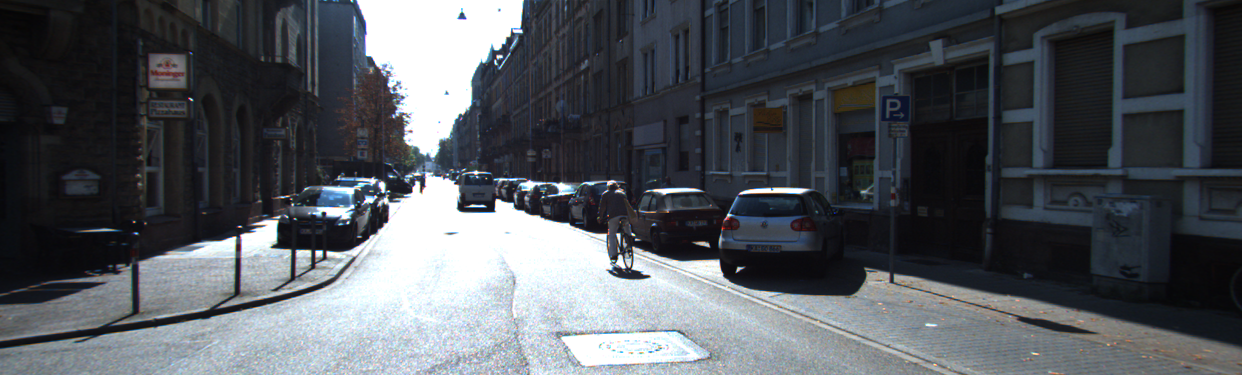}}
	
	\caption{Removing objects using 3D object detection in different stages. We use red, green and blue represent cars, cyclists and pedestrians respectively. Image in \cref{fig:corimage} is only utilized for visibility.}
	\label{fig:removeobjects}
	\vspace*{-5mm}
\end{figure}

These are the start points of the paper: localize in dynamic environments and detect full objects. The intuition is to directly sense dynamic objects and remove them. But when ego-robot moves, motion segmentation become implicit. However, this effect has little influence on object detection. When obtained objects previously, ego-motion estimation and moving objects detection in dynamic environments seem feasible.

In consideration of above all, we propose Dynamic Registration. The remainder of this paper is organized as follows. 
In \cref{relatedwork}, we review literature in related research areas. 
In \cref{methodology}, we describe the proposed Dynamic Registration.
In \cref{experiments}, we introduce the experimental details and evaluated results. 
Finally, we conclude this work and suggest future extensions in \cref{conclusions}.

%	\begin{figure}[h]
%		\vspace*{3mm}
%		\centering
%		\includegraphics[width=\columnwidth]{0000_pcd.png}
%		\caption{Results}
%		\label{fig:showcase}
%		\vspace*{-5mm}
%	\end{figure}

	\begin{figure*}[ht!]
		\begin{center}
			\includegraphics[width=170mm]{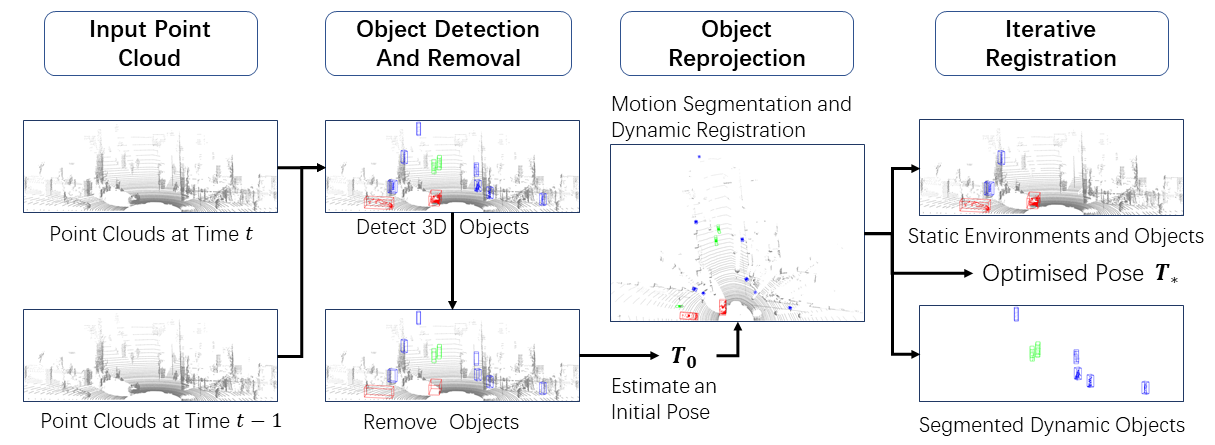}
			\caption{Over view of our Dynamic Registration system.}
			\label{fig:dr}
		\end{center}
	\end{figure*}

\section{Related Work}\label{relatedwork}
\subsection {Moving Object Detection}
Moving object detection or segmentation has been widely researched in the computer vision community in the past\cite{siam2018modnet}. Whereas here we only focus on lidar-based methods.
Generally, it's hard for MOD methods to segment complete moving objects in dynamic scenes due to the existing coupling of ego-motion, which threatens complete geometric representations of objects. 

\textit {Map based}:\cite{pagad2020robust} builds an occupancy map as a filter to remove dynamic points.\cite{pomerleau2014longterm} performs map points classification based on visibility assumptions for the sake of lifelong map updating and moving objects segmentation. 
\textit {Motion based}: Motion based methods detect moving objects without prior informations, but may fail to segment all moving objects limited by merely leveraging motion clues. And they are unable to detect stationary objects that have the potential to move\cite{dewan2016motionbased}.  
\textit {End to end}: Chen et al. \cite{chen2021moving} projects point cloud to range view and then generates residual images as inputs of a segmentation network. But it simply divided the point clouds into moving and static. 
\cite{sun2021pointmoseg} segments points using two secutive frames semantically.

However, these methods miss the concept of instance. They tend to segment a point with its state of motion and semantic label rather than to detect a car or a pedestrian with its size, location, orientation and, its motion state. That is to say, we may get some moving points representing a half car, which makes it hard to decide the motion state of the object.

\subsection {Localization in Dynamic Environment}
Most SLAM systems designed for dynamic environments are visual SLAM\cite{bescos2018dynaslam, yu2018dsslam}. 2D object detection is usually implemented in a robot SLAM system \cite{he2017mask}.
One important reason is, it's easier to deploy and accelerate a visual detection model than in lidar point clouds. However, recent advanced technics in 3D object detection make it possible to meet the need for both speed and accuracy in point cloud\cite{yan2018second,lang2019pointpillars}. We introduce them in \cref{section:3ddet}

Wang et al. \cite{wang2007simultaneous} first proposed a Bayesian-based theory of Simultaneous Localization, Mapping and Moving Object Tracking(SLAMMOT).
From then on, SLAM and the Detection And
Tracking of Moving Objects (DATMO) can be combined and have been proved to be beneficial to each other. Paralleling SLAM and MOT could estimate both ego-motion and object motion Simultaneously. 
\cite{moosmann2013joint} implement a joint estimation of localization and tracking. To emphasize the capability of tracking, they adopt segmentation rather than detection.
\cite{wang20204d} project 3D point cloud into a 2D plane, and adopt a FCN network to segment objects. But classification and clustering need to be taken as post-processing.

Our Dynamic Registration is designed for two point clouds frames without prior motion compensation or motion estimation, i.e., scene flow or GPS/IMU data. Once moving objects are segmented out, they can be tracked in any manner.

\section{Methodology}\label{methodology}
Our Dynamic Registration system consists of three modules, as illustrated in \cref{fig:dr}.
The first module performs a 3D object detection neural network, generating 3D bounding boxes for all instances.
The second module temporarily removes all objects from the input point cloud. And then, an initial pose can be drawn by classical scan match methods. 
In the third module, we transform objects in current frames into previous frames in Birds Eye View(BEV) by the estimated poses. And a motion state estimation module is employed to segment static and dynamic objects.
Points of static objects and environments merge to generate a new environment. Then the merged static environment is used to estimate the ego-pose and state of objects iteratively until no objects are considered static.
Finally, the static environment, ego-motion, and dynamic objects, these three coupling individuals, are calculated and estimated successfully.

\subsection{Object Detection And Removal}
\subsubsection{3D Object Detection}\label{section:3ddet}
We aim to remove moving object points from a registration pipeline and segment these points as instance-level objects. Previous works in segmenting or detecting moving elements use either a semantic segmentation or a binary moving segmentation since the heavy computations and complexities of instance segmentation in a 3D point cloud.

These 2D segmented methods are not enough to describe objects in 3D space. Further post-processing like clustering needs to be implemented in terms of instances.

The balance of real-time and accuracy need to be considered. 3D Object Detection predicts 3D bounding boxes for objects in a single frame point cloud. However, compared with 3D instance segmentation\cite{zhou2020joint}, 3D object detection achieves almost the same instance-level performances due to the sparsity of point clouds.

VoxelNet \cite{zhou2018voxelnet} achieve end-to-end detection using Voxel Feature Extractor, but introduced 3D convolutions. SECOND \cite{yan2018second} speeds up 3D convolutions via sparse optimization. PointPillars \cite{lang2019pointpillars} proposes pillar representation to improve the efficiency.
In order to implement real-time detections in scenes, we adopt PointPillars to realize 3D Object Detection.
The output of 3D object detection at time $t$ with $n$ objects is defined as
\begin{equation}\label{eq:3ddet}
	\begin{split}
		D_{t} &\triangleq \{ d_{0},d_{1},\ldots,d_{n} \}\\
		d_{n} &\triangleq \left\lbrace x,y,z,l,w,h,\theta,label,socre\right\rbrace 
	\end{split}
\end{equation}
where $ x,y,z $ the center point coordinate of a cuboid; $ l,w,h $ the length, width, height of the cuboid, both in meters, respectively; $\theta$ the heading angle around z-axis in degrees. In addition to these cuboid parameters, $label$ represents three classes including car, cyclist, and pedestrian, and $score$ denotes a confidence score for detection result.

\subsubsection{Removing Points In Cuboid}\label{section:points}

\begin{figure}[htbp]
	\vspace*{-3mm}
	\centering
	\includegraphics[width=\columnwidth]{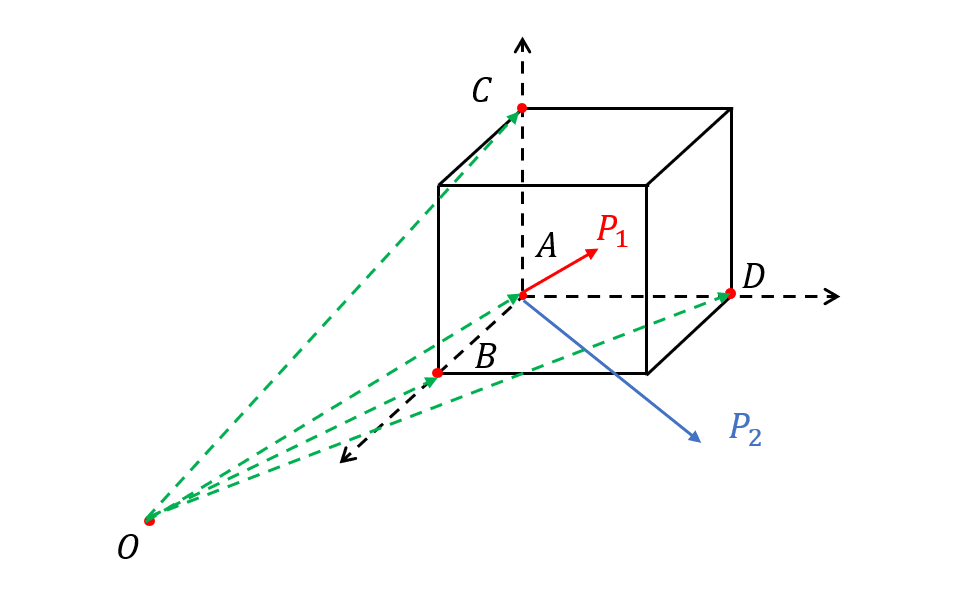}
	\caption{$ABCD$ represent a cuboid in a 3D space. O is origin of coordinates. $P_1$ (in red) is inside the cuboid while $P_2$ (in blue) is outside. $OA$, $OB$, $OC$, $OD$, $OP_1$, $OP_2$ denote their coordinates respectively.}
	\label{fig:findpoints}
    \vspace*{-3mm}
\end{figure}

Supposing a cuboid can be formulated as three vectors, as shown in \cref{fig:findpoints}, which are $AC$, $AB$, $AD$, respectively. We use the dot product to decide whether a point is located in the cuboid.

Let $P$ be a point in 3D space, if
\begin{equation}
	0 \leqslant \overrightarrow{AP} \cdot \dfrac{\overrightarrow{AB}}{\lVert \overrightarrow{AB} \rVert} \leqslant \overrightarrow{AB} \cdot \dfrac{\overrightarrow{AB}}{\lVert \overrightarrow{AB} \rVert}
\end{equation}

then $P$ is fixed at $\pm$ 90° around $\overrightarrow{AB}$ axis, and its length is within $\lVert \overrightarrow{AB} \rVert$. In the same way, the point $P$ can be limited in the cuboid. This procedure can be expressed as:

\begin{equation}
	\begin{split}
	0 &\leqslant \overrightarrow{AP} \cdot \dfrac{\overrightarrow{AB}}{\lVert \overrightarrow{AB} \rVert} \leqslant \overrightarrow{AB} \cdot \dfrac{\overrightarrow{AB}}{\lVert \overrightarrow{AB} \rVert}\\
	0 &\leqslant \overrightarrow{AP} \cdot \dfrac{\overrightarrow{AD}}{\lVert \overrightarrow{AD} \rVert} \leqslant \overrightarrow{AD} \cdot \dfrac{\overrightarrow{AD}}{\lVert \overrightarrow{AD} \rVert}\\
	0 &\leqslant \overrightarrow{AP} \cdot \dfrac{\overrightarrow{AC}}{\lVert \overrightarrow{AC} \rVert} \leqslant \overrightarrow{AC} \cdot \dfrac{\overrightarrow{AC}}{\lVert \overrightarrow{AC} \rVert}
	\end{split}
\end{equation}

We first remove all potential moving objects temporarily because static objects will be added to the environment after motion segmentation. \cref{fig:notremove} and \cref{fig:removeall} show the removing results.

\subsection{Object Reprojection}\label{section:objectreprojection}
\subsubsection{Initial Pose Estimation}\label{section:initpose}
We first remove all points in the detected cuboids without checking their motion state. After that, static and dynamic objects are both removed. Then an initial pose can be estimated by a registration algorithm using the clean point clouds.

\begin{equation}
	T_{0}=Registration(M_{t-1},M_{t})
\end{equation}
where $T_{0} \in SE(3)$.

\subsubsection{Reprojection}\label{section:reprojection}
Detection results $D_{t}$ at time $t$ is projected to time $t-1$ using estimated transformation matrix $T_0$. 

\begin{equation}
	D_{t}^{t-1}=T_{0}D_{t}
\end{equation}

Objects are detected differently in two frames. It theoretically remains the same detections in such a short time interval. However, affected by the uncertainty of observations and inferences, it may occur noisy detections, false positives, false negatives, and occlusions which result in different results for each frame.

\begin{figure}[htbp]
	\centering
	\includegraphics[width=\columnwidth]{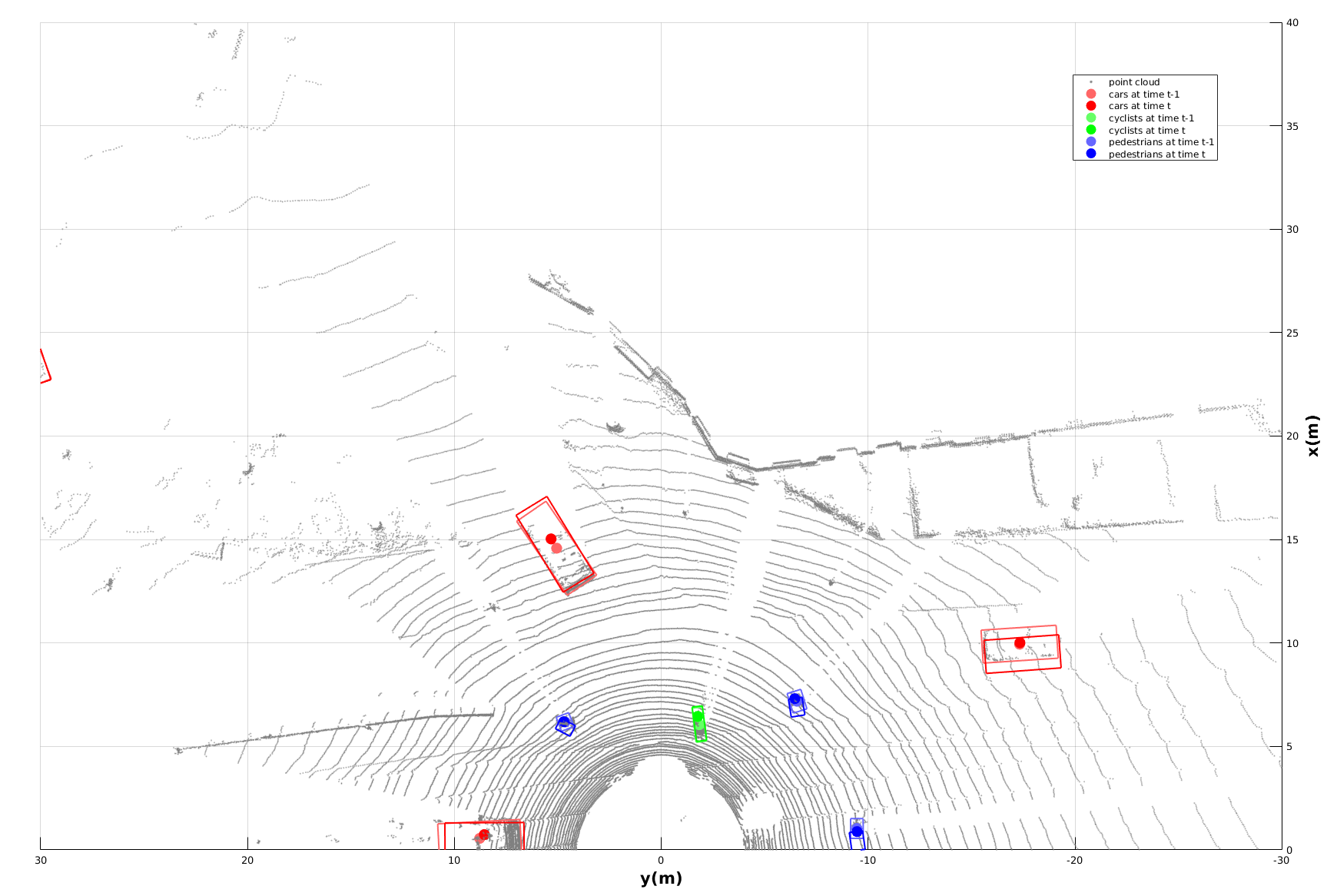}
	\caption{Object reprojection. Even objects are detected differently, they can be associated  and segmented.}
	\label{fig:reprojection}
	\vspace*{-3mm}
\end{figure}

\subsubsection{Data Association}\label{section:dataassociation}
Data association has to be taken to connect corresponding objects successfully.
Traditional methods include JPDA, PHD, MHT, RFS in the radar tracking system or deep learning based methods, but they might be too heavy for this situation. Inspired by SORT \cite{bewley2016simple} and AB3DMOT\cite{weng20203d} in the field of multi object tracking, here Hungarian assignment algorithm is employed for addressing the associate problem.

But different from the MOT task, objects have the closest distance when projected to the previous frame. Based on this fact, we can associate the center point of cuboids without calculating complex 3D IoU. Objects considered to be the same which has the nearest Euclidean distance, as shown in \cref{fig:reprojection}.

\subsection{Dynamic Registration}
\subsubsection{Motion segmentation}
Object reprojection errors can be drawn subsequently from associated detections according to the initial pose. Here the reprojection errors represent moving distances of objects in two frames, so detected objects can be segmented by a threshold. Then dynamic environment can be divided into moving objects and the static environment with static objects.

\begin{algorithm}
	\caption{Dynamic Registration}\label{algorithmDR}
	\KwIn{Input Point Clouds: $M_{t-1}$, $M_{t}$}
	\KwOut{Dynamic Registration Pose: $T^{DR}_{t}$ \\ 
		\qquad \quad \ \,  Segmented Dynamic Objects: $D_{t-1}^{d}$, $D_{t}^{d}$ \\
		\qquad \quad \ \,  Segmented Static Objects: $D_{t-1}^{s}$, $D_{t}^{s}$}
	
	$(D_{t-1},D_{t})\leftarrow $ 3D Object Detection $(M_{t-1},M_{t})$\\
	
	$(M_{t-1}^{RA}, M_{t}^{RA})\leftarrow $ RemoveAll $(M_{t-1}, D_{t-1}, M_{t}, D_{t})$\\
	
	$T_{t}^{0}\leftarrow $ Registration $(M_{t-1}^{RA}, M_{t}^{RA})$\\
	
	$D_{t}^{t-1}\leftarrow $ Object Reprojection $(T_{t}^{0}, D_{t-1}, D_{t})$\\
	
	$(D_{t-1}^{d}, D_{t}^{d}, D_{t-1}^{s}, D_{t}^{s})\leftarrow $ MotionSegmentation$(D_{t-1}, D_{t}^{t-1})$\\
	
	$(M_{t-1}^{RD}, M_{t}^{RD})\leftarrow $ MergeAndConcatenate$(D_{t-1}^{s}, D_{t}^{s}, M_{t-1}^{RA}, M_{t}^{RA})$\\
	
	$T^{DR}_{t}\leftarrow $ Registration $(M_{t-1}^{RD}, M_{t}^{RD})$\\
	
\end{algorithm}

\subsubsection{Iterative Dynamic Registration}
In most cases, removing all objects from point clouds helps improve registration accuracy. However, static objects always occur in highly changeling real-world environments, e.g., parked cars. Static objects also provide rich features. There is no direct intuition for keeping them from the registration pipeline. 

Since static objects have been segmented, they can be added to the environment. When newly stationary points are merging and concatenating, a registration algorithm, as mentioned in \cref{section:initpose}, is employed again for a more reliable transformation matrix. This dynamic registration process is iteratively conducted until no static objects are generated. After that, static environment, dynamic objects, and ego-motion are all fully segmented out. Any mapping, tracking, or localization algorithm may be utilized for post-processing.

We formulate the iterative dynamic registration process as \cref{algorithmDR}: Dynamic  Registration, \cref{algorithmIDR}: Iterative Dynamic Registration, respectively. Lower left $t$ represent time stamp.  Upper left $0$ and $\ast$ represent initial results and optimized results. Upper right symbols $RA$, $RD$, $s$, $d$ represent \underline{R}emove \underline{A}ll objects, \underline{R}emove \underline{D}ynamic objects, \underline{s}tatic and \underline{d}ynamic respectively.

%$D_{t-1}$,$D_{t}$ are detected 3D objects with inside points, in a format illustrated in \cref{eq:3ddet}; $T_{0}$ is the initial pose estimated from $M_{t-1}$ and $M_{t}$ utilizing traditional registration algorithm.

\begin{algorithm}
	\caption{Iterative Dynamic Registration}\label{algorithmIDR}
	\KwIn{Point Clouds: $M_{t-1}^{RA}$, $M_{t}^{RA}$\\
	\qquad \quad Detection Results:  $D_{t-1}$, $D_{t}$ \\
	\qquad \quad Initial Registration Pose: $T_{t}^{0}$ \\
	\qquad \quad Dynamic Registration Pose: $T_{t}^{DR}$}
	\KwOut{Iterated Registration Pose: $T^{\ast}_{t}$ \\ 
		\qquad \quad \ \,  Iterated Dynamic Objects: $D_{t-1}^{d\ast}$, $D_{t}^{d\ast}$ \\
		\qquad \quad \ \,  Iterated Static Objects: $D_{t-1}^{s\ast}$, $D_{t}^{s\ast}$}
	
	$^{0}D_{t}^{t-1}\leftarrow $ Object Reprojection $(T_{t}^{0}, D_{t-1}, D_{t})$\\

	 $(^{0}D_{t-1}^{d},\ ^{0}D_{t}^{d},\ ^{0}D_{t-1}^{s},\ ^{0}D_{t}^{s})\leftarrow $
	  Motion Segmentation $(D_{t-1},\ ^{0}D_{t}^{t-1})$\\

	$^{\ast}D_{t}^{t-1}\leftarrow $ Object Reprojection $(T^{DR}_{t}, D_{t-1}, D_{t})$\\
	
	$(^{\ast}D_{t-1}^{d},\ ^{\ast}D_{t}^{d},\ ^{\ast}D_{t-1}^{s},\ ^{\ast}D_{t}^{s})\leftarrow $ 
	Motion Segmentation $(D_{t-1}, ^{\ast}D_{t}^{t-1})$\\
	
	\While{$^{\ast}D_{t-1}^{s} \neq D_{t-1}^{s}$ and $^{\ast}D_{t}^{s} \neq D_{t}^{s}$}
	{$(^{\ast}M_{t-1}^{RD}, ^{\ast}M_{t}^{RD})\leftarrow $ MergeAndConcatenate$(^{\ast}D_{t-1}^{s}, ^{\ast}D_{t}^{s}, M_{t-1}^{RA}, M_{t}^{RA})$\\
		
	$T^{\ast}_{t}\leftarrow $ Registration $(^{\ast}M_{t-1}^{RD},^{\ast}M_{t}^{RD})$\\
	
 	$D_{t-1}^{s} = \, ^{\ast}D_{t-1}^{s}$ and $D_{t}^{s} = \, ^{\ast}D_{t}^{s} $ \\
 	
	$^{\ast}D_{t}^{t-1}\leftarrow $ Object Reprojection $(T^{\ast}_{t}, D_{t-1}, D_{t})$\\
	
	$(^{\ast}D_{t-1}^{d},\ ^{\ast}D_{t}^{d},\ ^{\ast}D_{t-1}^{s},\ ^{\ast}D_{t}^{s})\leftarrow $ MotionSegmentation$(D_{t-1}, ^{\ast}D_{t}^{t-1})$\\

}
	
\end{algorithm}

\section{Experiments And Results}\label{experiments}
To better evaluate the performance of our proposed dynamic registration algorithm, 
we adopt KITTI Tracking datasets \cite{geiger2013vision} as a benchmark for the experiment since it provides both ego and object information.

\subsection {Implementation Details}
As discussed in \cref{section:3ddet}, we leverage a pillars-based 3D object detection network, PointPillars, for detecting objects in KITTI Tracking sequences.

We directly use a version provided by MMDetection3D\cite{mmdet3d2020}, without any fine-tuning or editions. The model is pre-trained on KITTI 3D detection benchmark.  Detailed settings and the pre-trained model can be found in their repository.

3D object detection on KITTI dataset traditionally follows a range of [(0,70),(-40,40),(-3,1)]. However, it doesn't perform well in a registration scenario. Thus we  limit the point cloud range to [(0,40),(-30,30),(-3,1)] in x, y, z axis respectively. Subsequently, results out of range are filtered out. We also select cuboids with a confidence threshold higher than 0.5, in order to grab quality detections. 

Additionally, to improve the efficiency and accuracy of the registration algorithm, we downsample each input point cloud with a 0.3m voxel.

\subsection{Quantitative Experiments}

\subsubsection{Evaluation Metrics}

\begin{table*}[htbp]
	\centering
	\caption{Comparison of RMSE of RPE on KITTI Tracking Datasets Seq.0000-0020. Bold numbers indicate lower errors.}
	\begin{tabular}{c|cc|cc|cc}
		\toprule
		\multirow{2}[2]{*}{seq} & \multicolumn{2}{c|}{Registration} & \multicolumn{4}{c}{Dynamic Registration} \\
		& \qquad \quad NDT \qquad  &  \quad ICP \qquad \quad & \quad NDT-RMA \quad & \multicolumn{1}{c}{ICP-RMA} & \quad NDT-RMD \quad & \quad ICP-RMD \quad  \\
		\midrule
		0000  & 0.2583  & 0.3641  & 0.2646  & 0.3887  & \textbf{0.2101} & \textbf{0.3444} \\
		0001  & 0.3453  & 0.4550  & 0.3506  & 0.4793  & \textbf{0.2689} & \textbf{0.4050} \\
		0002  & 0.4560  & 0.5723  & 0.4718  & 0.5861  & \textbf{0.4479} & \textbf{0.5696} \\
		0003  & 0.7364  & 0.7952  & 0.7523  & 0.7869  & \textbf{0.7151} & \textbf{0.7769} \\
		0004  & 0.8349  & 0.7200  & 0.8509  & 0.7353  & \textbf{0.8127} & \textbf{0.6874} \\
		0005  & 0.8709  & 0.8537  & 0.8773  & 0.8643  & \textbf{0.8545} & \textbf{0.8301} \\
		0006  & 0.3247  & 0.4664  & 0.3190  & 0.4651  & \textbf{0.2912} & \textbf{0.4496} \\
		0007  & 0.3650  & 0.4428  & 0.3566  & 0.4601  & \textbf{0.3195} & \textbf{0.4095} \\
		0008  & 0.7415  & 0.7842  & 0.7361  & 0.7859  & \textbf{0.7102} & \textbf{0.7589} \\
		0009  & 0.4218  & 0.5042  & 0.4432  & 0.5230  & \textbf{0.3979} & \textbf{0.4613} \\
		0010  & 0.9602  & 0.8938  & \textbf{0.9185} & 0.8978  & 0.9407  & \textbf{0.8667} \\
		0011  & 0.3978  & 0.5209  & 0.4137  & 0.5629  & \textbf{0.3722} & \textbf{0.5111} \\
		0012  & 0.1161  & \textbf{0.1432} & \textbf{0.0856} & 0.1490  & 0.0955  & 0.1473  \\
		0013  & 0.3404  & 0.4283  & 0.3488  & 0.4282  & \textbf{0.2861} & \textbf{0.3735} \\
		0014  & \textbf{0.1874} & \textbf{0.4625} & 0.1939  & 0.5032  & 0.1876  & 0.4653  \\
		0015  & 0.3067  & 0.3625  & 0.3161  & 0.3705  & \textbf{0.2894}  & \textbf{0.3571} \\
		0016  & 0.1917  & \textbf{0.1136} & 0.1341  & 0.1145  & \textbf{0.1102} & 0.1159  \\
		0017  & 0.1259  & \textbf{0.1598} & 0.1514  & 0.1658  & \textbf{0.0801} & 0.1617  \\
		0018  & 0.2791  & 0.3746  & \textbf{0.2759}  & 0.3857  & 0.2788  & \textbf{0.3074} \\
		0019  & 0.2154  & 0.2977  & 0.2162  & 0.3054  & \textbf{0.1954} & \textbf{0.2797} \\
		0020  & 0.4879  & 0.7300  & 0.5098  & 0.7938  & \textbf{0.4526} & \textbf{0.7138} \\
		\midrule
		mean  & 0.4268  & 0.4974  & 0.4279  & 0.5120  & \textbf{0.3960} & \textbf{0.4758} \\
		\bottomrule
	\end{tabular}%
	\label{tab:addlabel}%
\end{table*}%

We adopt relative pose error (RPE) to test the performances of our proposed registration algorithm. Absolute trajectory error (ATE) may not be suitable in a scan matching scenario since a slight error in a transformation matrix at the beginning of a sequence leads to a large gap in two trajectories subsequently.
For simplicity, here we only consider the translational part of RPE: 
\begin{equation}\label{eq:rpe}
	\mathrm{RPE}_{\text{trans}}=\sqrt{\frac{1}{N-1} \sum_{i=1}^{N-1} \| \operatorname{trans}\left(\boldsymbol{T}_{\mathrm{gt}}^{-1}\boldsymbol{T}_{\text{est}} \right) \|_{2}^{2}}
\end{equation}
where $\boldsymbol{T}_{\mathrm{gt}}$ and  $\boldsymbol{T}_{\text{est}}$ are ground truth pose and estimated pose between two frames, respectively.

\subsubsection{Results}

The results are shown in \cref{tab:addlabel}, which details comparisons of our method’s performance against other traditional registration algorithms. We build our Dynamic Registration based on two classical methods, NDT and ICP. Results match intuitive expectations on average. Dynamic Registration performs better in almost all sequences, since removing dynamic objects follows the static world assumption. 

Results show our approach outperforms other registration algorithms, even better than removing all potential moving objects.
Removing all objects without checking their motion state also works when objects all keep moving. However, this coarse strategy increases localization error because fewer static points are engaged to the registration pipeline. Adding static points to the environment decreases localization uncertainty in a dynamic and complex environment.

\subsection{Qualitative Evaluation}
\subsubsection{Odometry And Mapping}

\begin{figure}[htbp]
	\centering
	\subfigure[traj]{
		\label{fig:traj}\includegraphics[width=\columnwidth]{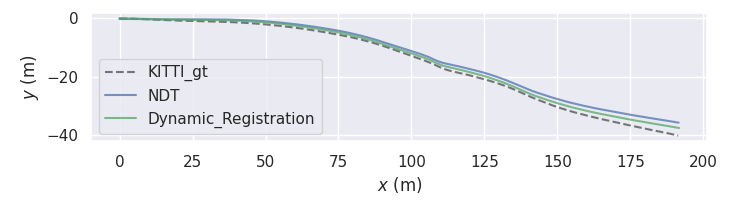}}
	
	\subfigure[Odometry and mapping results of NDT]{
		\label{fig:ndtmapping}\includegraphics[width=0.9\columnwidth]{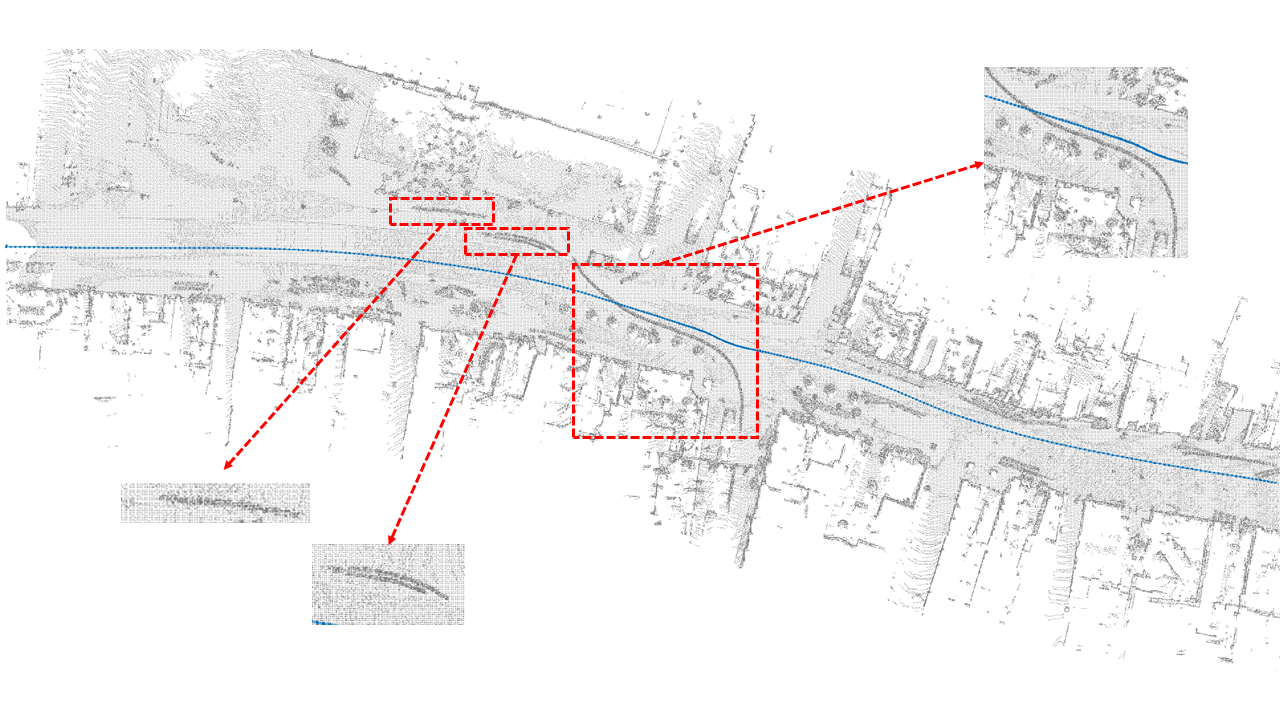}}
	
	\subfigure[Odometry and mapping results of Dynamic Registration ]{
		\label{fig:rmdmapping}\includegraphics[width=0.9\columnwidth]{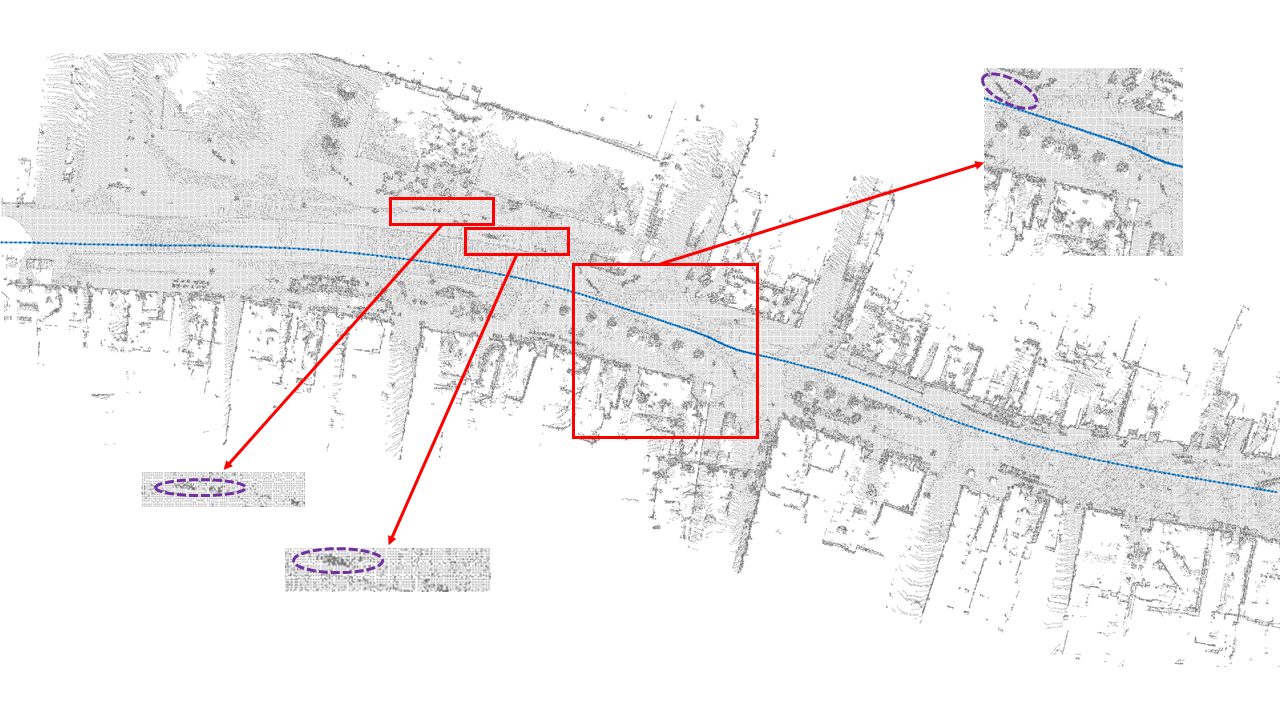}}
	
	\caption{Comparisons of the global trajectories for sample sequence 0013 among KITTI Tracking Datasets. In \cref{fig:traj}, dashed curves represent ground truth poses. Blue solid lines are trajectories of NDT, and the red ones are trajectories of proposed Dynamic Registration. In \cref{fig:ndtmapping} and \cref{fig:rmdmapping}, red box zoom-in details o f the environment.}
	\label{fig:odoandmap}
	\vspace*{-5mm}
\end{figure}

We also implement a pairwise structureless odometry based on continuous point cloud registration simply on visualization grounds. Since other factor makes it inexplicit to evaluate the effects of removing dynamic objects. Results are shown in \cref{fig:odoandmap}. Here Dynamic Registration is build on NDT, corresponding to NDT-RMD in \cref{tab:addlabel}. Dynamic Registration Odometry is closer to the ground truth trajectory than NDT odometry. 

Mapping results indicate that our approach successfully removes dynamic objects. Red boxes enlarge objects' trajectories. Obvious ghostings are cleared, but fragment remains(purple box in \cref{fig:rmdmapping}), mainly due to false negatives of the detector. However, false positives matter little, since they are possibly segmented as static objects and then added to the environment.
 
\subsubsection{Moving Object Detection}

In most cases, moving objects can be segmented successfully. But motion segmentation in dynamic registration relates to 3D object detection, ego-motion estimation, and the preset segmentation threshold.
Low-quality detection results occur when objects are too far from the ego car, which leads to false positives or false negatives. These "wrong detections" are filtered out by data association, as mentioned in \cref{section:reprojection}. But data association merely solved the issue partially, since false negatives remain until state estimation and prediction are conducted. Detect objects in point cloud sequences (also regarded as multi object tracking) is beyond the discussion scope of this paper.

\section{Conclusions}\label{conclusions}

In this paper, we proposed dynamic registration, combining ego motion estimation and moving object detection in a point cloud registration pipeline. 3D object detection is utilized to detect objects previously and removed temporarily. Then it begins to iteratively estimate ego motion and add static objects until no static objects are segmented. We evaluate our proposed method on KITTI benchmark and compare it with different registration algorithms, which demonstrates its effectiveness. We'll future extend dynamic registration to a dynamic slam system.

\bibliographystyle{IEEEtran}
\bibliography{reference/cite}

\end{document}